%% file: acl_latex.tex
\documentclass[11pt]{article}

\usepackage[preprint]{acl}

\usepackage{times}
\usepackage{latexsym}
\usepackage[T1]{fontenc}
\usepackage[utf8]{inputenc}
\usepackage{microtype}
\usepackage{inconsolata}
\usepackage{graphicx}

\usepackage{amsmath}
\usepackage{amssymb}
\usepackage{booktabs}
\usepackage{needspace}
\usepackage{xcolor}

\newcommand{\blfootnote}[1]{%
  \begingroup
  \renewcommand\thefootnote{}\footnote{#1}%
  \addtocounter{footnote}{-1}%
  \endgroup
}

\newcommand{\dsseven}{DeepSeek-R1-Distill-Qwen-7B}
\newcommand{\qweneight}{Qwen3-8B}
\newcommand{\pp}{\,\text{pp}}
\newcommand{\simplek}{simple@32}

\title{Stable Answers, Unfinished Reasoning: Why Self-Consensus\\ Is Not a Safe Early-Exit Signal}

\author{
  Yunxiang Mo\thanks{\ Equal contribution.} \\
  HKUST \\
  \texttt{ymoaj@connect.ust.hk}
  \And
  Donghao Zhao\footnotemark[1] \\
  HKUST \\
  \texttt{dzhaoah@connect.ust.hk}
  \And
  Hejia Geng \\
  University of Oxford \\
  \texttt{hejia@tapntell.ai}
}

\begin{document}
\maketitle
\blfootnote{Code and data will be released at
\url{https://github.com/Antony-zdh/stable-answers-unfinished-reasoning}.}

\input{00_abstract}
\input{01_introduction}      
\input{02_related_work}      
\input{04_method}            
\input{06_mechanism}         
\input{05_results}           
\input{09_conclusion}        

\input{10_limitations}       

\bibliography{custom}

\appendix
\input{A_appendix}                 
\input{G_self_consensus}           
\input{H_practice}                 

\end{document}

%% file: 00_abstract.tex
\begin{abstract}
A natural way to cut reasoning-model inference cost is to repeatedly probe a
single partial trajectory for its current answer and stop once probes
\emph{agree}---\emph{self-consensus}. We ask whether any such rule is both safe
and token-saving, and whether one can be selected once and reused. A
\emph{preregistered} sweep of $3{,}520$ consensus rules, replayed on frozen
trajectories from two models and three benchmarks, clears none of three
acceptance gates fixed in advance; the frontier reproduces on a held-out split
and on two unseen models---while a boundary-confidence control
(DEER) swept through the same pipeline clears all three. The reason lies in the
signal: agreement establishes that the
current answer \emph{persists} under a fixed probing procedure, not that the
reasoning has \emph{terminated}---a \emph{consensus--termination gap}. Stopping
on it commits non-terminal answers. At a rule still saving $32\%$ of the
tokens, one stop in nine fires on an answer the trajectory itself later
abandons, and most of those stops cut off a correction it would otherwise have
made. Widening the agreement window does not remove them: the share levels off
near $7\%$, and by then the saving has fallen to $8\%$. Probe re-wording and a
hand-labelled error taxonomy show the agreed answer is often a
placeholder the model had not settled on. Used on its own as the stop signal,
agreement fails not because it is insufficiently strict, but because it
repeatedly measures the wrong object.
\end{abstract}

%% file: 01_introduction.tex
\section{Introduction}
\label{sec:intro}

Reasoning language models \citep{deepseekr1,qwen3,openai_o1} can spend thousands
of tokens on a single answer. \emph{Early exit}---halting a chain of thought once
its reasoning is done---could cut that cost \citep{overthinking,dynasor_certaindex},
but it requires knowing \emph{when the reasoning has finished}, which the
trajectory does not announce. A natural and increasingly studied proxy is
\emph{self-consensus}: repeatedly probing a \emph{single} partial trajectory for
its current answer (e.g., by appending a short ``\texttt{Final Answer:}''
suffix) and stopping once recent probe answers \emph{agree}
\citep{dynasor_certaindex}. We abbreviate self-consensus as \emph{consensus}
below, and reserve \emph{self-consistency} for the distinct setting in which
several trajectories are sampled independently (\S\ref{sec:related}).
The premise is that if the model
repeatedly says $x$, it has probably finished reasoning about $x$; recent work
reports answers that ``converge'' well before a trajectory ends and stops on that
convergence at little \emph{aggregate} accuracy cost
\citep{liu2025answerconvergence}.

We show that this proxy measures the wrong object. Under a fixed probing
procedure, repeated agreement establishes only that the current answer
\emph{persists}, not that the reasoning has \emph{terminated}, and the two come
apart---a \emph{consensus--termination gap}. This explains the aggregate savings
reported above. They are real, but they are measured without charging for the
probe tokens they spend, and the accuracy they cost is \emph{directional}: a
stop far more often locks in a wrong answer the trajectory would have corrected
than it saves a right one.

The gap has a concrete source. A probe requires an answer even when the
underlying trajectory has not settled, so repetition under a fixed query often
records the elicitation procedure rather than reasoning termination; the
elicited placeholder can repeat across many probes and look settled while the
trajectory goes on to correct
it. Three direct measurements bear this out (Section~\ref{sec:mechanism}).
Re-probing the \emph{same} frozen prefix with a differently worded query returns a
different answer $54\%$ of the time in the first tenth of a trajectory against
$16\%$ near the end, so early ``answers'' are largely artifacts of being asked.
Hand-labelling $134$ wrong stops finds more than half commit to an answer the model
had not converged on, against one in four it had genuinely settled. And a
consensus stop frequently commits a non-terminal answer: at a rule still saving
$32\%$ of the tokens, one stop in nine commits an answer the trajectory later
leaves behind, and of the $216$ such stops $155$ pre-empt a correction the
trajectory went on to make against $16$ that bank a correct answer
from a trajectory ending wrong. The computation a consensus stop removes is
predominantly the computation that would have corrected the answer, the opposite
of what early exit is for. Widening the agreement window does not close the
gap. The non-terminal share flattens near $7\%$ while the saving that pays for
it collapses to $8\%$, and the accuracy loss shrinks only because the rule
fires less often and later. Compared at a matched saving, the accuracy price stays
high: at $28\%$ net saving the cheapest of the $3{,}520$ rules below costs
$6.0\pp$ and at $33\%$ it costs $8.3\pp$, against $0.5$ and $2.0\pp$ for DEER,
the non-consensus control below (\S\ref{sec:mech-ratio}).

Prior reports establish good accuracy--saving points for a particular model,
benchmark and parameter setting. We ask two stricter questions: whether
\emph{any} rule in the family is simultaneously safe and saving at all, and
whether one can be \emph{selected} once and reused rather than retuned per model
and benchmark. To answer both, and to rule out a badly chosen threshold or a
single implementation, we search the signal exhaustively under commitment. On frozen trajectories from two
models and three benchmarks we run a \emph{preregistered} sweep of the windowed
consensus family we study---window size $\times$ share threshold plus operational
knobs, $3{,}520$ rules---with problem-grouped splits and acceptance gates fixed in
advance (Section~\ref{sec:results}). No setting of these knobs clears any gate:
\textbf{not one of the $3{,}520$ rules is both safe and saving} on development,
the split we select on, so no rule is ever carried forward. On train, $478$
rules clear a gate in-sample, enough for a study reporting on one split to have
concluded otherwise. The frontier reproduces on the test
split ($r{=}0.98$) and on unseen models at $4\times$ the scale and a different
architecture. A shipped operating point, CertaIndex (CoT), loses
$56$--$70\pp$ on the same trajectories.

\begin{figure*}[t]
  \centering
  \includegraphics[width=\textwidth]{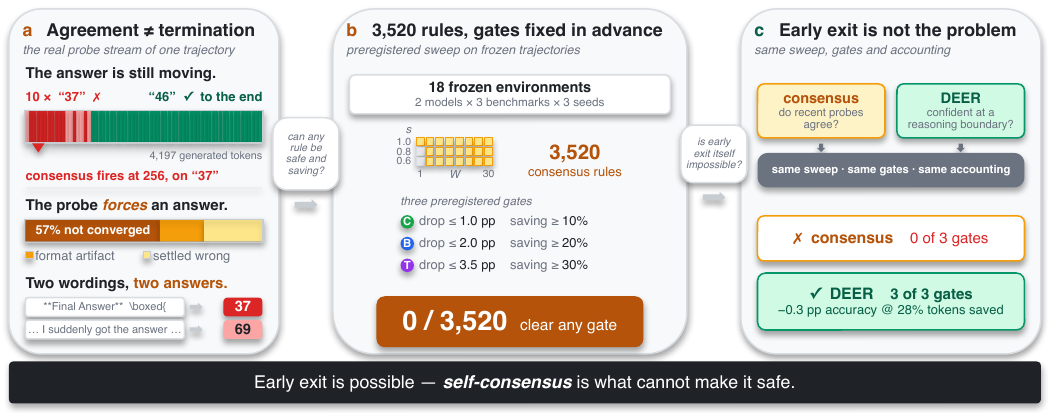}
  \caption{\textbf{The argument in one picture.}
  \textbf{(a)} On \textsc{MATH}500 problem~68 (\dsseven{}, dense $64$-token
  probes) the model emits the same wrong answer, \texttt{37}, for ten
  consecutive probes; a three-probe rule fires at token~$256$ and commits it.
  Left to run, the trajectory reaches the correct \texttt{46}. Ribbon colours:
  red the dominant wrong answer, pink other wrong answers, green correct.
  Below it, the bar splits the $134$ hand-labelled wrong stops of one
  environment (\S\ref{sec:mech-placeholder}), and the two rows re-probe the
  \emph{same} frozen prefix at token~$256$ with our answer suffix and with the
  CertaIndex suffix (\S\ref{sec:mech-wording}).
  \textbf{(b)} The preregistered sweep of $3{,}520$ rules, replayed on $18$
  frozen environments against three gates fixed in advance. None clears any
  gate.
  \textbf{(c)} Swept through the same pipeline, gates and accounting, a
  boundary-confidence method (DEER) clears all three---on the test split, at
  $4\times$ scale and on a different architecture too (\S\ref{sec:exp-heldout}).}
  \label{fig:idea}
\end{figure*}

Early exit itself is not the obstacle. Swept through the same
pipeline and gates, DEER---a non-consensus method that reads the model's
confidence at a reasoning boundary---clears all three, losing as little as
$0.3\pp$ while saving $28\%$, on the unseen models too
(\S\ref{sec:exp-deer}; Figure~\ref{fig:idea}).

\needspace{5\baselineskip}
\paragraph{Contribution.}
\begin{itemize}
  \item \textbf{We identify a consensus--termination gap in probe-based early
  exit}: under a fixed probing procedure repeated agreement establishes only
  that the current answer persists, not that the reasoning has terminated
  (Section~\ref{sec:mechanism}).
  \item \textbf{We explain what agreement measures}, through three direct
  signals---the wording sensitivity of the extracted answer, a hand-labelled error
  taxonomy of stopped-but-wrong cases, and the non-terminal
  answers a consensus stop commits---which together show that early agreement is a
  probe-elicited placeholder on an unfinished trajectory rather than a settled
  belief (Section~\ref{sec:mechanism}).
  \item \textbf{We show the gap is not a threshold or implementation artifact}: a
  preregistered sweep of $3{,}520$ rules on frozen trajectories closes no gate,
  while a non-consensus signal swept under the same protocol clears the same
  gates (Section~\ref{sec:results}). We will release the protocol,
  trajectories, and sweep in full upon publication.
\end{itemize}

%% file: 02_related_work.tex
\section{Related Work}
\label{sec:related}

\paragraph{Probe-based / consensus early exit.}
The methods we scrutinize periodically query a partial chain of thought
\citep{wei2022cot} for its current answer and halt on agreement. Dynasor / CertaIndex
\citep{dynasor_certaindex} formalize this with a ``certainty index'' over recent
probe answers and use it to terminate early when serving reasoning programs, a
response to the \emph{overthinking} of reasoning models
\citep{overthinking,deer}. Recent work pushes the same idea further:
\citet{liu2025answerconvergence} report that a trajectory's answer ``converges''
well before it ends and stop on that convergence at little \emph{aggregate}
accuracy cost. We adopt the same probe mechanism (a short
boxed-answer suffix) and notion of self-consensus, and ask what those results
leave open. They establish good accuracy--saving points for a given model,
benchmark and parameter setting; we ask whether the family contains one rule
that is safe, saving, and transferable across benchmarks, models and seeds
without retuning.

\paragraph{Cost-aware comparisons of exit policies.}
Closest to our question, \citet{dong2026learnstop} compare a learned
prefix-feature stopper against calibrated confidence, entropy and
running-answer exits across $18$ task--model settings, at a matched tolerance
for lost correct answers. Of those baselines, running-answer exits are the
closest analogue to the signal we study. They find three regimes: learning wins
on free-form mathematics, calibrated scalar exits win on multiple choice, and
the hardest benchmarks admit no certifiable aggressive policy at all. That last regime is a
partial precedent for our negative result. Two things separate the two. Their
matched tolerance is far looser than our $1.0\pp$ conservative gate, so
``learned stopping wins'' there and ``no rule of $3{,}520$ clears a gate'' here
are distant points on one trade-off curve. And
where they compare policies, we dissect one signal, asking why agreement
carries no termination information and answering with a mechanism
(\S\ref{sec:mechanism}).

\paragraph{Gating agreement behind a reasoning-level signal.}
\citet{min2026puma} reach the same diagnosis from the design side: answer-level
signals, they argue, reflect answer \emph{readiness} rather than reasoning
convergence, and can fire while the model is still exploring or correcting
itself. Their remedy, PUMA, retains answer agreement but consults it only once a
detector finds successive reasoning steps semantically redundant, so the
reasoning---not the answer---decides when a stop is even considered. Across
five models and five benchmarks it holds accuracy while removing about a
quarter of the tokens, where their reproduction of a pure consensus stopper
\citep{liu2025answerconvergence} loses $23$--$59\pp$ at $82$--$91\%$ token
reduction. That gap is the directional cost of \S\ref{sec:mech-ratio}, on an
independent implementation. The questions differ. They take the gap as a
design premise and build a remedy; we measure it (\S\ref{sec:mechanism}) and ask
whether \emph{any} setting of the consensus signal escapes it, against gates
fixed in advance. And their design keeps agreement as one component gated by
another signal, where our negative result concerns agreement used
\emph{alone}---the scope \S\ref{sec:locating} states, and a reason to read the
two together.

\paragraph{Self-consistency and independent sampling.}
Self-consistency \citep{wang2023selfconsistency} samples a \emph{diverse} set of
reasoning paths and takes the majority answer; agreement is evidence there
because the paths are drawn independently, so the vote can outweigh an error on
any single path, and early-stopping variants only decide how many samples to
draw \citep{aggarwal2023adaptive,li2024esc}. Probing a single chain of
thought---\emph{self-consensus} (\S\ref{sec:intro})---is a different object:
each probe reads a longer prefix of the one path the earlier probes read, so
agreement records that
one line of reasoning has not changed its answer, not that separate lines of
reasoning have converged---the protecting vote is absent. On its own, though, this
structural point does not settle whether agreement tracks \emph{termination}.
The corrections a stop pre-empts (\S\ref{sec:mech-ratio}) are ones the
trajectory makes as it continues, not revisions solicited after a completed
answer, which models perform poorly without external feedback
\citep{huang2024selfcorrect}.

\paragraph{Alternative termination signals.}
A stop can read signals other than consensus: verifier- and
process-reward-guided decoding \citep{lightman2024verify,wang2024mathshepherd}
scores partial reasoning, adaptive-computation methods learn a halting signal
at the token or layer level \citep{graves2016act,schuster2022confident},
confidence-dynamics stops read how the model's answer confidence evolves
\citep{hosseini2026codestop}, building on the finding that a model's own
confidence carries usable signal \citep{kadavath2022know}, and DEER
\citep{deer} stops on the model's
confidence in a trial answer at reasoning boundaries. We sweep DEER through our
own pipeline in Section~\ref{sec:exp-deer}. We do not claim this broader class
is categorically superior; our experiments isolate one signal.

\paragraph{Test-time budget control.}
A different axis dispenses with detection and fixes the compute from outside;
test-time compute is a resource that trades against accuracy whether or not a
signal decides how much to spend
\citep{snell2024scaling,brown2024monkeys}. Budget forcing
\citep{muennighoff2025s1} suppresses the end-of-thinking token to lengthen
reasoning, and length-controlled reinforcement learning \citep{aggarwal2025l1}
trains a model to respect a requested length. Budget forcing is evidence from
the opposite side for the mechanism of \S\ref{sec:mechanism}: forcing a
trajectory to continue past the point where it tries to conclude raises AIME24
accuracy from $50$ to $57\%$ \citep{muennighoff2025s1}, so a trajectory that
presents a finished answer need not have finished. Length control sets the
operating point on the trade-off
surface of \S\ref{sec:exp-sweep} by fiat; we ask whether a signal can find it
automatically.

%% file: 04_method.tex
\section{Experimental Setup}
\label{sec:method}

Every measurement in this paper---the direct measurements of the gap
(Section~\ref{sec:mechanism}) and the rule sweep that tries to tune it away
(Section~\ref{sec:results})---runs on the same frozen substrate under the same
token accounting, so that no result can be an artifact of a rule changing the
reasoning it observes.

\subsection{Frozen Trajectories, Offline Probes}
For each problem we generate exactly \emph{one} main trajectory in a single
request (caps: $16$K tokens for \textsc{MATH}500 and AMC23, $32$K for AIME24)
and freeze it. All probing is then performed \emph{offline} against fixed text
prefixes of that frozen trajectory. Because probes never re-enter the decode,
changing a probe schedule---or a stopping rule---cannot change the underlying
reasoning; every rule is scored against the same trajectories. We build two
offline probe banks: a \emph{dense} bank that re-probes every $64$ tokens with
\simplek{} (a boxed-answer probe capped at $32$ output tokens), and an
\emph{adaptive} bank that
additionally fires event-triggered probes at entropy drops and at
conclusion/reflection/answer markers.

\subsection{Token Accounting}
\label{sec:accounting}
For a rule that stops a problem at token position $s$ after consuming probe
output $p$, we charge total decode tokens $T=s+p$ and compare against the frozen
baseline $B$ (the full trajectory, capped at budget). \emph{Net} savings is
$(B-T)/B$ and \emph{gross} savings is $(B-s)/B$; their gap is the probe cost.
Probe \emph{prefill} is not charged: $p$ counts decode only. This follows
\citet{dynasor_certaindex} and \citet{deer}, both of which reuse the main
branch's KV cache for probe prompts. The choice is also the more conservative
one here, since both of those works additionally exclude probe \emph{output} from
their reported budgets, which we charge
(Appendix~\ref{sec:appendix-tables}).
The accuracy drop compares the correctness of the committed answer at $s$
against that of the frozen final answer; it is not an agreement rate between the
two.

\subsection{Models, Benchmarks, Splits}
\label{sec:resolution}
Development uses two models---\dsseven{} and \qweneight{}---on three benchmarks
(\textsc{MATH}500, AMC23, AIME24), each at seeds $42/43/44$: $18$
model$\times$benchmark$\times$seed environments. Problems are partitioned
\emph{by problem id} into $60/20/20$ train/dev/test splits, so no problem leaks
across splits. Confirmation reserves seeds $45/46/47$ plus two unseen
models---\mbox{DeepSeek-R1-Distill-Llama-8B} and \mbox{Distill-Qwen-32B}---all
evaluated on the \emph{test} split only, read once after all rules and
thresholds are frozen. Every headline metric is a
\emph{macro}-average over model$\times$benchmark$\times$seed environments,
weighting each equally, so that the largest benchmark does not dominate policy
selection (Appendix~\ref{sec:appendix-prereg} details the weighting and its
noise trade-offs).

%% file: 06_mechanism.tex
\section{The Consensus--Termination Gap}
\label{sec:mechanism}

This section establishes the paper's central object by direct measurement.
The source of the gap is that a probe elicits an
answer rather than reading one: what repeats across probes can be a placeholder
that looks stable while the trajectory keeps correcting itself. We establish, in
order: agreement is pervasive and non-terminal (\S\ref{sec:exp-fc}); the early
answer depends on how the probe is worded (\S\ref{sec:mech-wording}); most
stopped-but-wrong commitments are to answers the model had not converged on
(\S\ref{sec:mech-placeholder}); no amount of required agreement removes the
non-terminal stops (\S\ref{sec:mech-ratio}); and non-independent readings supply no vote that could
catch the error (\S\ref{sec:mech-independence}).

\subsection{The First Actionable Consensus}
\label{sec:exp-fc}
An online rule cannot wait for a trajectory to end: it fires the first
time its condition is met, so what matters is what the \emph{first actionable
consensus} commits to. On a dense probe stream logged for every \textsc{MATH}500
problem under \dsseven{} at the main $16$K/$32$K budgets ($500$ problems $\times$
three seeds $=1{,}500$ trajectories; Appendix~\ref{sec:appendix-fc}), we take
that to be the first point at which
three consecutive probes agree, are non-empty, and carry no hedging marker such
as ``wait'' or ``but''.

\textbf{(1) The first actionable consensus is usually not terminal.} It fires on
$1{,}477/1{,}500$ trajectories, and the answer it commits is correct only
$50.5\%$ of the time, whereas the same trajectories run to completion reach
$90.7\%$: a $40.2\pp$ loss.

\textbf{(2) End-of-trajectory agreement is a different object.} Once a
trajectory finishes, agreement does track correctness ($97.8\%$ for cumulative
agreement across all probes, $90.4\%$ for the final five-probe window), but no
such end-state evidence is available mid-trajectory. Recovery separates the
two: $736$ of the $1{,}500$ trajectories at some point agreed on an answer they
went on to abandon, and $84.2\%$ of those still ended correct.

\paragraph{The same question on a broader environment set.}
Facts (1)--(2) read one model on one benchmark. We therefore repeat the
measurement across all $18$ development environments (both models, three
benchmarks, three seeds, dev split; $684$ trajectories), this time dropping the
hedging-marker filter so that the rule fires as early as it can. First consensus
forms in $679$ of the $684$ trajectories, half of them within the first tenth of
the trajectory's own length. There the committed answer is correct only $27.5\%$
of the time against $85.2\%$ run to completion. Consensus that forms later is
more often right---$77\%$ for the few that first agree between $40$ and $60\%$
of the way through (Figure~\ref{fig:consensus-pos}, appendix)---so waiting for a
later consensus trades saving for accuracy.

\subsection{Probe Wording Versus Position}
\label{sec:mech-wording}
If early agreement reflected a settled belief, the answer a probe returns should
not depend much on how the probe is worded. We test this directly on paired
re-probes of the \emph{same frozen prefixes}: at every probe position we read the
trajectory with two different probe wordings (our boxed-answer suffix and the
CertaIndex suffix, both capped at $32$ tokens), so the model's state is identical
and only the elicitation differs. We read all $18$ development environments, on
the $652$ of $684$ trajectories that finish inside the budget; for a truncated
trajectory, position as a fraction of its own length is undefined.

\begin{figure*}[t]
  \centering
  \includegraphics[width=\textwidth]{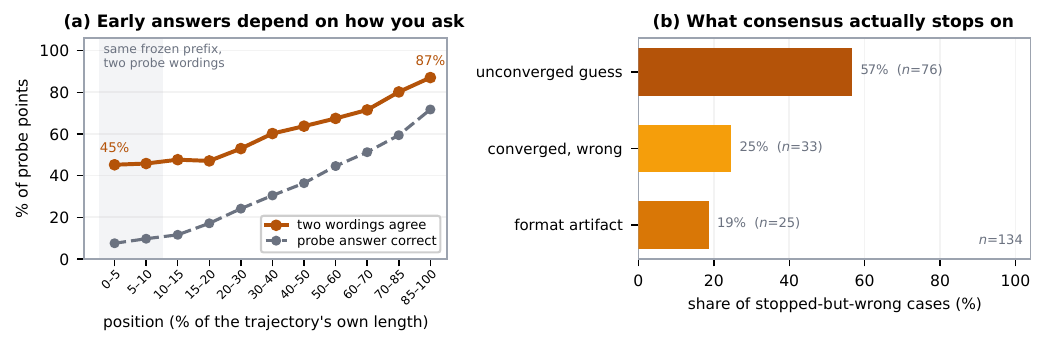}
  \caption{\textbf{Two sides of the consensus--termination gap.}
  \textbf{(a)} The same frozen prefix (all $18$ development environments)
  re-probed with two differently worded answer queries. Early on the two
  wordings return the \emph{same} answer less than half the time ($45\%$ in the
  earliest bin, $46\%$ over the first tenth); by the end they usually coincide
  ($87\%$ in the last bin).
  Position is a fraction of each trajectory's own length.
  \textbf{(b)} All $134$ stopped-but-wrong cases, labelled independently by two
  annotators (Appendix~\ref{sec:appendix-repro}): $56.7\%$ commit to an
  answer the model had \emph{not converged} on, $18.7\%$ to a probe-format
  artifact, $24.6\%$ to a genuinely settled wrong value.}
  \label{fig:wording-taxonomy}
\end{figure*}

Figure~\ref{fig:wording-taxonomy}(a) shows the result. In the first tenth of a
trajectory the two wordings return different answers $54\%$ of the time; by the
final third they disagree only $16\%$ of the time. A stable but mistaken belief
would have both wordings read back the same wrong value, so an early probe is
not reading back an answer the trajectory holds but forcing one out of reasoning
that has not converged.

Across all $46{,}360$ paired
positions the two wordings read back answers that are almost equally often
correct (a $0.9\pp$ difference), so the early sensitivity is about \emph{which}
answer is elicited, not a defect of one suffix.

\subsection{Error Taxonomy}
\label{sec:mech-placeholder}
Two annotators independently labelled \emph{all} $134$ cases in which probes
agreed on a wrong answer, from an exploratory single-environment pass
(\dsseven{}, \textsc{MATH}500, seed $42$, $3{,}072$-token cap;
Appendix~\ref{sec:appendix-repro}). This pass is diagnostic: nothing in
\S\ref{sec:results} rests on it. Only $24.6\%$ are cases where
the model had genuinely settled on a wrong value
(Figure~\ref{fig:wording-taxonomy}(b)). $56.7\%$ are answers it had not
converged on at all, and a further $18.7\%$ are artifacts of the probe's output
format.

A consensus rule cannot tell a forced placeholder apart from a settled belief.
The same account fits the window effect of \S\ref{sec:mech-ratio}: a longer
window is a longer buffer in which a placeholder can be overturned, so it
screens out some non-terminal agreement, though without removing the underlying
risk.

\subsection{The Cost of a Non-Terminal Stop}
\label{sec:mech-ratio}
The natural remedy is to demand more agreement before stopping---a stop only
when the last $W$ probes agree, for larger $W$. It helps, but the help has to be
bought with the saving. We measure the cost on all $3{,}420$ development-model trajectories
(train, dev and test), with one rule per $W$ and $W$ the only axis varying, in
two steps.

\begin{figure}[t]
  \centering
  \includegraphics[width=\columnwidth]{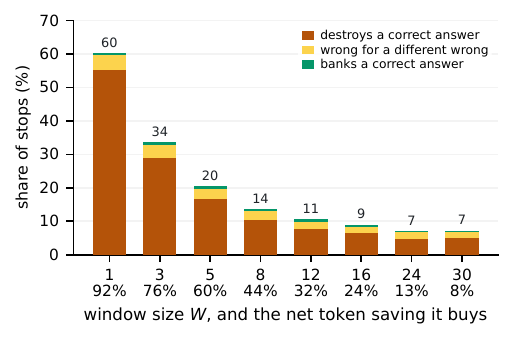}
  \caption{\textbf{The non-terminal share falls with $W$, and so does the
  saving that pays for it.} Share of stops that commit an answer the trajectory
  later leaves behind, split by what the change costs; the annotation above each
  bar is that total share. It falls from $60\%$ to a floor near $7\%$ while the
  net saving falls from $92\%$ to $8\%$ (\S\ref{sec:mech-ratio}). All $3{,}420$
  development-model trajectories; one rule per $W$ (share $1.0$, interval
  $128$, maturity $512$, schema validity).}
  \label{fig:nonterminal-stops}
\end{figure}

\textbf{Is the committed answer terminal?} Often not. At $W{=}12$, a rule still
saving $32\%$ of the tokens, $10.6\%$ of stops commit an
answer the trajectory later leaves behind; at the short windows that save most,
the share reaches $60\%$ ($W{=}1$; Figure~\ref{fig:nonterminal-stops}). Demanding
more agreement does drive the share down, from $60\%$ at $W{=}1$ to $10.6\%$ at
$W{=}12$ and $7.0\%$ at $W{=}24$, but the saving is what pays for it, falling
from $92\%$ to $13\%$ over that same range. And the share does not reach zero;
past $W{=}24$ it stops falling at all ($7.2\%$ at $W{=}30$, for $8\%$ saving).
\textbf{Is that a loss?} Mostly. Of the $216$ non-terminal stops at
$W{=}12$, $155$ commit a wrong answer to a trajectory that would have reached
the right one, $45$ trade one wrong answer for another, and $16$ bank a correct
answer from a trajectory that ends wrong. The computation consensus removes is
disproportionately the computation that fixes the answer.

The exchange is unfavourable throughout. DEER (the non-consensus control of
\S\ref{sec:exp-deer}), by contrast, buys the same saving far more cheaply: at $28\%$ net saving it costs
$0.5\pp$ of accuracy where the cheapest of our rules costs $6.0\pp$, and at
$33\%$, $2.0$ against $8.3\pp$ (training and development splits together,
$2{,}736$ trajectories),
because its boundary-confidence signal commits
mainly once the trajectory has settled.

These counts are restricted to trajectories that finish inside the budget, since
a truncated one has no final answer to compare against. Counting a truncated
trajectory as incorrect instead, so that any correct stop on one is a rescue,
leaves the picture unchanged wherever the rule still saves ($155{:}16$ becomes
$155{:}57$ at $W{=}12$) and reverses it only at $W{=}30$, which saves $8\%$
($30{:}39$; Appendix~\ref{sec:appendix-tables}).

Appendix~\ref{sec:appendix-cases} sets three probe streams side by side, among
them a first-probe placeholder held for $27$ consecutive probes, longer than
all but the largest window.

\subsection{The Role of Independence}
\label{sec:mech-independence}
The readings a consensus rule aggregates are not independent, so agreement among
them cannot protect accuracy the way independent sampling does.
Self-consistency's
majority vote can absorb an error because its paths are sampled independently
(\S\ref{sec:related}), so a wrong answer on one path is outvoted by others that
reasoned differently. Probing one trajectory yields no such thing. Successive
probes observe nested prefixes from the same evolving reasoning path and
therefore share most of their reasoning history, so the readings move together,
and when that shared history is wrong they are wrong together. Agreement across
probes is thus a statement about one line of reasoning not having changed its
answer, not corroboration from several.

%% file: 05_results.tex
\section{Searching the Consensus Rule Space}
\label{sec:results}

Section~\ref{sec:mechanism} measures the gap. The natural response is that a
better-tuned rule avoids it---a wider window, a stricter share, a maturity
floor, a certainty requirement. This section rules that out by exhaustive
search under commitment: a \emph{preregistered} sweep of the consensus family
with acceptance gates fixed in advance \citep{preregistration_ml}, a
non-consensus signal swept through the same pipeline, a released
consensus stopper at its shipped default, and out-of-distribution confirmation.
Two questions run through it. First, whether \emph{any} rule in the family is
simultaneously safe and saving on the split the gates are applied to. Second,
if one were, whether it would be an \emph{environment-robust policy}---selected
once rather than tuned per environment, and still safe and saving on held-out
problems, seeds, benchmarks and models.

\subsection{Preregistered Rule Space and Gates}
\label{sec:schema}
The consensus signal in the windowed family we study reduces to two
hyperparameters: a \textbf{window size} $W$ (how many of the most recent probes
to inspect) and a \textbf{share threshold} $s$ (what fraction of that window
must agree on a single answer before a stop is allowed). $W{=}1$ is the
latest-probe rule; $s{=}1.0$ is strict unanimity over the last $W$ probes;
an entropy-over-the-window criterion closely tracks the
same agreement and adds no materially new operating point on our grid. We sweep
$W\in\{1,3,5,8,12,16,24,30\}$---deliberately including large windows, so
a rule can demand a long, stable agreement---and $s\in\{0.6,0.8,1.0\}$, crossed
with the operational knobs that govern \emph{when} and \emph{how} the signal is
read: probe schedule (four fixed intervals or event-triggered probing), a
minimum-token maturity floor, an answer-shape validity filter, and an optional
certainty flag. Enumerating this preregistered grid yields $3{,}520$ candidate
consensus rules (all values and formal definitions in
Appendix~\ref{sec:appendix-schema}).

\paragraph{A non-consensus control.}
To separate a failure of the consensus family from a failure of early exit
itself, we sweep one non-consensus method through the same pipeline, gates, and
token accounting. \textbf{DEER} \citep{deer} stops on the model's own confidence
in a trial answer at a reasoning boundary; its only hyperparameter is a
confidence threshold, swept over $14$ values (the stronger
\emph{trial-answer-submit} variant). Both are placed on one frontier under
identical token accounting. It is a control on the family, not a one-factor
ablation of the statistic (\S\ref{sec:locating}).

\paragraph{Gates.}
We fix three operating points and their acceptance gates \emph{before} looking
at held-out data, applied in order: the total accuracy drop (macro over the
$18$ environments) must stay under a cap, then the total net token saving must
clear a floor, and finally net saving must be positive on a minimum fraction of
environments (psf).

\begin{table}[t]
  \centering\small
  \begin{tabular}{lccc}
    \toprule
    Operating point & Max total & Min total & psf \\
                    & acc.\ drop & saving    & floor \\
    \midrule
    conservative     & $1.0\pp$ & $10\%$ & $0.80$ \\
    balanced         & $2.0\pp$ & $20\%$ & $0.80$ \\
    token\_efficient & $3.5\pp$ & $30\%$ & $0.70$ \\
    \bottomrule
  \end{tabular}
  \caption{Preregistered acceptance gates, fixed in advance and \emph{not}
  relaxed post hoc. All three conditions are macro-averaged over the $18$
  development environments; psf is the positive-saving fraction, the share of
  environments on which net saving must be positive.}
  \label{tab:gates}
\end{table}

The three points (Table~\ref{tab:gates}) pair progressively looser accuracy
constraints with progressively higher saving requirements. They are not
universal safety standards; they test whether self-consensus supplies a usable
policy under \emph{some} practically meaningful pairing of the two. The saving
floor rejects a rule that stays accurate only by almost never stopping, and the
psf floor an average saving concentrated in a few environments. Three commitments were recorded
with the protocol: the test split and confirmation models are never read
during selection (C1); the gates are never relaxed post hoc---an empty gate is
reported as the finding (C2); and the rule set is frozen and hashed before any
test evaluation (C3). Appendix~\ref{sec:appendix-prereg} gives the
preregistration text.

\subsection{Gate Outcomes Across the Rule Space}
\label{sec:exp-sweep}
We replay all $3{,}520$ rules on the frozen trajectories and aggregate to
per-environment metrics---$126{,}720$ rows $=3{,}520$ rules $\times$ $18$
environments $\times$ $2$ splits (train, dev)---then apply the gates on dev.
\textbf{Not one of the $3{,}520$ rules clears any gate.} The outcome holds
across environments: accuracy strictly falls in $64.9\%$ of all
rule--environment evaluations and strictly rises in only $10.5\%$, for a mean
drop of $13.0\pp$. The reason is the shape of the trade-off, read directly off
the sweep's own two axes (Figure~\ref{fig:ws-heatmap}, appendix): only $40$
rules keep the total drop at or below the conservative $1.0\pp$ cap, and the
most any of them saves is $0.2\%$---far under the $10\%$ floor. Conversely, the
first rule to save $10\%$ already costs $2.66\pp$, saving $20\%$ costs
$6.17\pp$, and saving $30\%$ costs $11.8\pp$. The size of that margin is
specific to the protocol's macro averaging: pooled over problems the gates are
likewise empty, but the best safe rule falls only just short of the saving
floor rather than nowhere near it
(Appendix~\ref{sec:appendix-tables}). The large windows behave exactly
as \S\ref{sec:mech-ratio} predicts: they do buy low drop, but only by stopping
so late that net saving collapses to zero, and no cell of the $(W,s)$ surface
clears the conservative gate. Nor is the result an artifact of charging dense
re-probing: forgiving the \emph{probe tax} entirely and applying the same three
gates to \emph{gross} saving still clears no rule, and the safe rules' gross
saving reaches only $2.07\%$. The tax opens a $4$--$7\pp$ gap between gross and
net saving and is reducible by a sparser schedule; the accuracy floor the gates
bind on is not
(Appendix~\ref{sec:appendix-tables}).

\subsection{A Non-Consensus Control Under the Same Gates}
\label{sec:exp-deer}
The same gates are cleared by a rule that does not read consensus. Sweeping DEER's confidence threshold through the same
pipeline and accounting, its trial-answer-submit variant passes all three
operating points on dev: the conservative point loses only $0.33\pp$ while
saving $28.2\%$, the balanced point $1.03\pp$ at $29.6\%$, and the
token-efficient point $2.75\pp$ at $31.9\%$; at a stricter threshold it is
accuracy-neutral ($+0.06\pp$) while still saving $20.8\%$
(Table~\ref{tab:main}; full threshold frontier in Appendix~\ref{sec:appendix-tables}). Early exit is therefore not impossible at these
budgets; the consensus \emph{signal} is what cannot make it safe.

\begin{table}[t]
  \centering\small
  \setlength{\tabcolsep}{3pt}
  \begin{tabular}{lccc}
    \toprule
    Method & Acc. & $\Delta$acc & Net \\
           &      & (macro)     & saving \\
    \midrule
    Full generation (consensus) & $82.5\%$ & --- & $0\%$ \\
    Consensus (drop$\le1.0$)  & $81.6\%$ & $-0.9\pp$  & $+0.2\%$ \\
    Consensus (save$\ge10\%$) & $79.9\%$ & $-2.7\pp$  & $+10.9\%$ \\
    \midrule
    Full generation (DEER)   & $82.9\%$ & --- & $0\%$ \\
    DEER (conservative)      & $82.5\%$ & $-0.33\pp$ & $+28.2\%$ \\
    DEER (balanced)          & $81.9\%$ & $-1.03\pp$ & $+29.6\%$ \\
    DEER (token\_eff.)       & $80.1\%$ & $-2.75\pp$ & $+31.9\%$ \\
    \bottomrule
  \end{tabular}
  \caption{Dev operating points (macro over $18$ environments). Each
  $\Delta$acc is against the baseline printed above it, which its own replay
  reproduces environment by environment; the $0.39\pp$ between the two
  baselines is a grading difference on a single problem plus a completion-gate
  convention, and moves no gate
  (Appendix~\ref{sec:appendix-repro}).}
  \label{tab:main}
\end{table}

\subsection{Consensus in the Wild}
\label{sec:exp-prior}
The sweep covers rules we constructed; a released consensus stopper shows where
a shipped default lands. \textbf{CertaIndex (CoT)} \citep{dynasor_certaindex},
reproduced from the released implementation at its default setting---stop once
three consecutive probes give the same non-empty answer and none hedges, with
the authors' own probe prompt and schedule---stops on $98.7$--$99.8\%$ of our
frozen trajectories and loses $56$--$70\pp$ while saving $77$--$90\%$ of
tokens. A faithful DEER \emph{readout} reproduction on
the same trajectories stays near full accuracy ($+0.8\pp$ on \qweneight{}).
Table~\ref{tab:baselines} (appendix) reports both reproductions with scope
notes; in particular, nothing here bears on CertaIndex's multi-path setting,
which resembles self-consistency over independently sampled trajectories.

\subsection{Generalization}
\label{sec:exp-heldout}
We read the test split once (commitment C3); with no selected rule to confirm,
we ask whether the frontier \emph{reproduces}. It does: each rule's total drop
on the held-out test split tracks its dev value at $r{=}0.98$. On the two unseen
models---\mbox{Distill-Qwen-32B} ($4\times$ the development scale) and
\mbox{DeepSeek-R1-Distill-Llama-8B} (a different architecture and family),
three test seeds each---the frontier reproduces at $r{=}0.97$ and $r{=}0.94$,
and the conservative gate stays empty on both. DEER, swept identically,
clears the gates on dev, on test and on both unseen models. On the $32$B it
gains $0.24\pp$ of accuracy while saving $32.4\%$, and on Llama it loses
$0.67\pp$ at $26.7\%$. Appendix~\ref{sec:appendix-pareto} traces the selection pipeline
end to end (Figure~\ref{fig:split-transfer}), gives per-model and per-benchmark
frontiers with an oracle upper bound, and details the dev/test asymmetry
induced by the small competition sets.

\subsection{Locating the Failure}
\label{sec:locating}
Every method that stops on \emph{consensus}---the swept family and the
CertaIndex (CoT) reproduction alike---fails a gate or loses many points, while
the one method that clears them does not read consensus. Run through the
same machinery, gates, and accounting, the two outcomes differ, so the failure
lies in how the stop is decided.

The DEER contrast changes two things at once, \emph{where} a stop is
considered and \emph{what statistic} decides it, so we vary them one at a time
(Table~\ref{tab:ablation}). On the savings axis both matter.
Moving the read positions from our probe grid to DEER's reasoning boundaries
lifts the best safe saving from $0.2\%$ to $3.9\%$, and moving the statistic at
our own positions lifts it to $13.9\%$. The $28.1\%$ total separates into
$13.7\%$ from the statistic, $8.9\%$ from the positions and $5.5\%$ from the
cost of our denser probing.

Only the statistic moves the accuracy axis, which is what the gates turn on.
Asked for the conservative gate's $10\%$ saving, the cheapest
consensus rule loses $2.7\pp$ at our positions and $3.8\pp$ at DEER's; asked for
the balanced gate's $20\%$, $6.2\pp$ and $10.2\pp$. Reading DEER's confidence at
our own positions instead gains $0.7\pp$ at the conservative floor and
costs $3.1\pp$ at the balanced one. Under gross accounting, which forgives the
probe output $p$ on each side, it gains at both.

The two factors do different work. A schedule sets which stop positions $s$ a
rule can reach and how much probe output it must spend to reach them, so moving
it acts on the probe cost, the gap between gross and net saving. The statistic
decides whether the reasoning at such an $s$ has settled, so it governs the
accuracy the stop costs and how early an $s$ can be committed, which is
gross saving. That is why only changing the statistic lowers the accuracy price
at every floor. The claim is about agreement used \emph{alone}. Combined with a
signal that reflects the model's own estimate, the cheap fact that the answer
has not changed may still be useful, provided the combination avoids the risk
measured here.

\begin{table}[t]
  \centering\small
  \setlength{\tabcolsep}{4pt}
  \begin{tabular}{llcrr}
    \toprule
    \multicolumn{2}{c}{Stop rule} & Best safe & \multicolumn{2}{c}{$\Delta$acc at} \\
    \cmidrule(r){1-2}\cmidrule(l){4-5}
    Statistic & Reads at & saving & $10\%$ & $20\%$ \\
    \midrule
    Agreement  & our grid   & $0.2\%$ \,(0/3)  & $-2.7$ & $-6.2$ \\
    Agreement  & boundaries & $3.9\%$ \,(0/3)  & $-3.8$ & $-10.2$ \\
    Confidence & our grid   & $13.9\%$ (0/3)   & $+0.7$ & $-3.1$ \\
    Confidence & boundaries & $28.3\%$ (3/3)   & $+0.1$ & $+0.1$ \\
    \midrule
    \multicolumn{5}{l}{\emph{probe cost forgiven on both sides (gross)}} \\
    Agreement  & our grid   & $2.1\%$ \,(0/3)  & $-1.9$ & $-3.9$ \\
    Agreement  & boundaries & $5.4\%$ \,(0/3)  & $-2.7$ & $-9.6$ \\
    Confidence & our grid   & $21.5\%$ (2/3)   & $+0.7$ & $+0.7$ \\
    Confidence & boundaries & $30.4\%$ (3/3)   & $+0.1$ & $+0.1$ \\
    \bottomrule
  \end{tabular}
  \caption{Varying the two factors that separate the swept family from DEER,
  one at a time, on the $659$ development problems for which DEER records at
  least one reasoning boundary (same grader, same gates, macro over the $18$
  environments). ``Best safe saving'' is the largest saving among rules losing
  at most $1.0\pp$, with the number of gates cleared in parentheses.
  ``$\Delta$acc at $10/20\%$'' is the best $\Delta$acc at which a cell can buy
  that much saving ($+$ a gain, $-$ a loss)---the accuracy price of a fixed
  budget. The lower block repeats both under gross accounting, which forgives each rule's
  own probe cost. Appendix~\ref{sec:appendix-tables} gives the full frontier.}
  \label{tab:ablation}
\end{table}

%% file: 09_conclusion.tex
\section{Conclusion}
\label{sec:conclusion}

Used on its own, self-consensus is not a safe early-exit signal: under a fixed
probing procedure agreement establishes that the current answer \emph{persists},
not that the reasoning has \emph{terminated}---a \emph{consensus--termination
gap}. What the probes agree on is often a placeholder forced from an unfinished
trajectory, so a stop commits an answer the trajectory itself goes on to
abandon, pre-empting the correction that would have followed. Demanding more
agreement reduces such stops only by giving the saving back. That gap leaves the
safe-and-saving corner empty across $3{,}520$ preregistered rules, while a
boundary-confidence control swept through the same protocol clears every gate.
The failure is not that agreement is too easy to reach, but that reaching it is
no evidence of termination.

%% file: 10_limitations.tex
\section*{Limitations}
\label{sec:limitations}

\paragraph{Evidence for the mechanism.}
Our account of \emph{why} consensus fails---that probes of one trajectory are not
independent readings, and that they frequently record a forced placeholder---rests
on a structural argument (Sections~\ref{sec:related}
and~\ref{sec:mech-independence}) together with the paired wording experiment,
the hand-labelled error taxonomy and the window sweep. We do not run a controlled
experiment that manipulates independence directly, and the labelled cases were
collected under a short window, the regime where unconverged answers are most
common. A stronger test would construct probe streams with varying degrees of
dependence and measure how the accuracy tax responds.

\paragraph{Scope of the negative result.}
The central result---no consensus rule clears any gate---is established on $18$
development environments (two models, three benchmarks, seeds $42/43/44$) and
confirmed on the held-out test split and two unseen models
(Section~\ref{sec:results}). The result is a statement about the space searched: one
boxed-answer probe suffix (\simplek{}) and one preregistered consensus schema
($3{,}520$ rules over a window-size $\times$ share-threshold family plus
operational knobs). Other probe wordings and signals outside this schema are not
searched; the mechanism (Section~\ref{sec:mechanism}) is our argument that
consensus variants would not help, but it is an argument, not an exhaustive
proof.

\paragraph{Scope of the DEER comparison.}
We sweep DEER's trial-answer-submit variant to establish that \emph{some} signal
clears the gates the consensus family fails (on the development models, the
held-out test split, and both unseen models); we do not claim it is the best
possible early-exit method, and we do not evaluate it as a leaderboard entry.

\paragraph{Small held-out sets on hard benchmarks.}
AIME24 and AMC23 are small (dev splits of $6$ and $8$ problems per seed), so
those per-environment cells are noisy.
The total drop macro-averages over the $18$ development environments (three seeds
per benchmark$\times$model), and we read the unseen models as
frontier-reproduction evidence rather than as independent gate tests
(Section~\ref{sec:resolution}).

\paragraph{Probe-tax dependence of the savings axis.}
Net-savings numbers charge dense re-probing (every $64$ tokens, $32$-token
probes); a sparser or KV-cache-reusing schedule would change the \emph{savings}
axis. We separate gross from net savings (Section~\ref{sec:accounting}) so the
probe-tax-dependent part is distinguishable, and \S\ref{sec:exp-sweep} reports
the gates applied to gross saving, where the outcome is unchanged; the
accuracy-tax result does not depend on the schedule.

\paragraph{Domain.}
All benchmarks are competition mathematics with checkable final answers. Whether
the consensus--termination gap and the accuracy tax transfer to open-ended reasoning, code, or
agentic tasks is out of scope.

%% file: A_appendix.tex
\section{Rule Schema Details}
\label{sec:appendix-schema}

The consensus schema of Section~\ref{sec:schema} centers on two hyperparameters
and a few operational knobs, enumerating to $3{,}520$ candidate rules over a
fixed-schedule and an event-triggered family. Each rule fixes:
\begin{enumerate}
  \item \textbf{Window size} $W\in\{1,3,5,8,12,16,24,30\}$: the number of most
  recent probes inspected. $W{=}1$ is the latest-probe rule.
  \item \textbf{Share threshold} $s\in\{0.6,0.8,1.0\}$: the fraction of the last
  $W$ probes that must agree on one answer (share measured over the full window,
  so an empty or dissenting probe counts against consensus); $s{=}1.0$ is strict
  unanimity. An entropy-over-the-window criterion is not swept separately: on our
  grid it tracks the same agreement closely enough to add no materially new
  operating point.
  \item \textbf{Probe schedule}: fixed interval $\in\{64,128,256,512\}$ tokens,
  or \texttt{adaptive\_event} with entropy-drop and marker triggers and a
  fallback interval.
  \item \textbf{Maturity}: a minimum-token floor $\in\{0,512,1024,2048,4096\}$
  before any stop is allowed.
  \item \textbf{Certainty}: optionally require the supporting probes flagged
  certain (minimum fraction).
  \item \textbf{Validity}: answer-shape filter (reject empty / single-letter
  answers to suppress the Type-E probe artifact of
  Appendix~\ref{sec:appendix-fc}), or accept any non-empty answer.
\end{enumerate}

\paragraph{Certainty.}
A probe is \emph{certain} when its output carries no hedging marker
(``wait'', ``hold'', ``but'', ``okay'', ``no'', ``hmm''). Dimension~5 gates a
stop on the fraction of the supporting probes that are certain.

\paragraph{Entropy triggers.}
The \texttt{adaptive\_event} schedule fires an extra probe when the token-level
predictive entropy of the main decode drops by at least $\Delta H$ relative to a
trailing baseline (an ``the model just committed'' signal), and at
conclusion/reflection/answer marker tokens (e.g., ``therefore'', ``the answer
is''), subject to a minimum inter-probe gap and a fallback interval.

\section{Preregistration Text}
\label{sec:appendix-prereg}

The protocol version, split manifest hash, candidate-rule generation script, and
the three operating-point gates (Table~\ref{tab:gates}) were fixed before any
held-out evaluation. \emph{Preregistered} here means fixed in the repository's
protocol file and hashed before the evaluation it governs, not deposited with a
third-party registry. The commitments of Section~\ref{sec:schema} in full:
\textbf{(C1)} the test split and both confirmation models are never read during
rule selection. \textbf{(C2)} All three operating points and their gates are
fixed before seeing held-out data and are not relaxed afterwards; if no rule
passes a gate, that is reported as the finding rather than engineered away, and
all three points are reported regardless of outcome. \textbf{(C3)} The frozen
rule set is fixed on train+dev and its hash recorded before any
test/confirmation evaluation. The conservative gate proved empty, and we report
it as such for both consensus and DEER.

\paragraph{Metric resolution.}
The total accuracy drop is a macro-average over the $18$ environments (each
model$\times$benchmark$\times$seed weighted equally). Pooling problems instead
would let \textsc{MATH}500 ($100$ dev problems per seed vs.\ $8$ for AMC23 and
$6$ for AIME24) dominate the headline, so a rule that wins only on
\textsc{MATH}500 could pass; equal per-environment weight forces a selected
rule to hold on the harder competition sets too; three seeds per cell damp the
per-environment noise. The gates are read on
the dev split (the selection target) with train reported alongside; where the
text says ``train+dev,'' a rule was required to pass a gate in-sample on train
and is then measured on dev.

\section{Reproducibility}
\label{sec:appendix-repro}

\begin{figure*}[t]
  \centering
  \includegraphics[width=\textwidth]{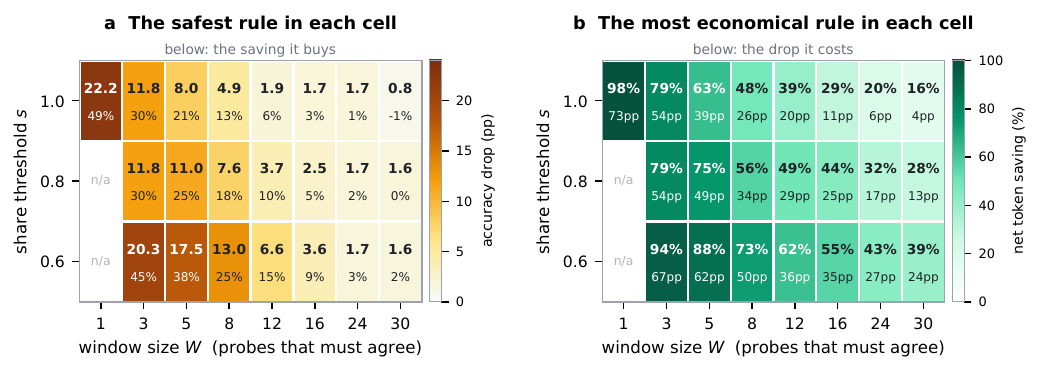}
  \caption{\textbf{The safe-and-saving corner is empty across the entire rule
  space.} Dev split, macro-averaged over the $18$ environments. Each cell holds
  the $160$ rules sharing its $(W,s)$, summarised by the two members a gate
  interrogates: \textbf{(a)} the cell's \emph{safest} rule (colour: accuracy
  drop; number: the saving it buys) and \textbf{(b)} its \emph{most economical}
  rule (colour: net saving; number: the drop it costs). A cell would be outlined
  in green if any of its rules cleared the conservative gate (drop
  $\leq 1.0\pp$ \emph{and} saving $\geq 10\%$); none is.}
  \label{fig:ws-heatmap}
\end{figure*}

Frozen trajectories, offline probe banks, the sweep archive
($126{,}720$ train+dev metric rows plus the DEER threshold sweep), and the
aggregation scripts will be released upon publication.
Table~\ref{tab:deer} and Figure~\ref{fig:ws-heatmap} were reproduced to the
reported precision directly from the sweep archive. Grading uses a robust
answer-equivalence check that
tries the raw and normalized reference forms and both string and
math-expression equality; per-problem correctness for the frozen baseline is
recomputed at replay time rather than trusted from collection.

\paragraph{Grading dependencies.} The robust check needs
the released \texttt{dynasor} evaluator, whose grading stack has to be pinned;
the released \texttt{pyproject.toml} pins it, so \texttt{pip install -e .}
installs a working one. The \texttt{antlr4-python3-runtime==4.7.2} pin is
load-bearing, since \texttt{latex2sympy2} raises
\texttt{Could not deserialize ATN} against newer runtimes. A missing evaluator
is the more dangerous case: the replay code carries a numeric fallback that
would grade $0.5$ against $\tfrac{1}{2}$ as wrong, so an incomplete install
returned systematically low accuracy with no warning. The grader now refuses to
run in that state rather than degrading, and
\texttt{scripts/grader\_selfcheck.py} confirms a working one: it checks
that the two forms above compare equal, and that full-generation accuracy,
regraded from the frozen development trajectories rather than read from the
stored flags, macro-averages to $82.94\%$ over the $18$ environments.

\paragraph{Two full-generation baselines.} That $82.94\%$ is a live regrade, and
it sits $0.39\pp$ above the $82.55\%$ Table~\ref{tab:main} prints for full
generation under the consensus sweep, the reference
that table's consensus rows, and every accuracy drop in
\S\ref{sec:exp-sweep}, are measured against. Two differences
account for it exactly; applying both to the live regrade reproduces the
archive's accuracy in each of the $18$ environments separately, not merely on
average (\texttt{scripts/baseline\_reconciliation.py}). $0.33\pp$ is one
\textsc{MATH}500 problem whose answer differs from the reference only in the
order of two summands: the frozen archive grades it wrong and the current
grader grades it right, and because the same problem is in all six
\textsc{MATH}500 development cells, one problem moves six environments by a
point each. The remaining $0.06\pp$ is a scoping choice rather than a grading
one: the replay scores a baseline trajectory that did not finish inside its
budget as wrong whatever answer it had reached, whereas the self-check grades
the stored answer either way. Exactly one development trajectory changes
verdict under it. The same two terms place the table's second baseline, the one
its DEER rows are measured against, between the other two: the DEER bank applies
the budget gate and grades that one problem right, which puts it at $82.88\%$.

Neither difference can move a result. Grading does not enter the token
accounting, so the discrepancy is confined to the accuracy axis, and every gate
pairs an accuracy cap with a saving floor. On that axis the cheapest rule in the
family that reaches even the $10\%$ conservative floor costs $2.66\pp$ against a
$1.0\pp$ cap, a margin more than four times the discrepancy, while the best
rule inside the cap saves $0.21\%$ against that floor.

\paragraph{Coding of the error taxonomy.} The three classes of
\S\ref{sec:mech-placeholder} are a simplification of the scheme the annotators
actually worked with, which had five labels: a collapse onto a wrong
\textbf{numeric} value (A), onto a wrong closed-form \textbf{expression} (B), a
\textbf{sign or symbol} error (C), a \textbf{reasoning gap} (D), where the probe
read back an intermediate quantity, a placeholder or a guess emitted before any
final candidate had formed, and a \textbf{format artifact} (E), such as an
option letter returned for a free-response item. The finer split was there
to keep the labelling concrete, but A, B and C are one thing seen three ways:
the trajectory had converged, on a wrong answer. We therefore pool them into
the single \emph{converged, wrong} class the paper reports, leaving D and E as
they stand. Two annotators labelled all $134$ cases independently against the
five-way scheme; the per-case labels ship at the five-way granularity, so any
other grouping can be recounted. Where the two disagreed, the released files
will carry both annotators' labels alongside the label of record and the ground
on which that case was decided, so the resolution is inspectable case by case.

\begin{figure*}[t]
  \centering
  \includegraphics[width=\textwidth]{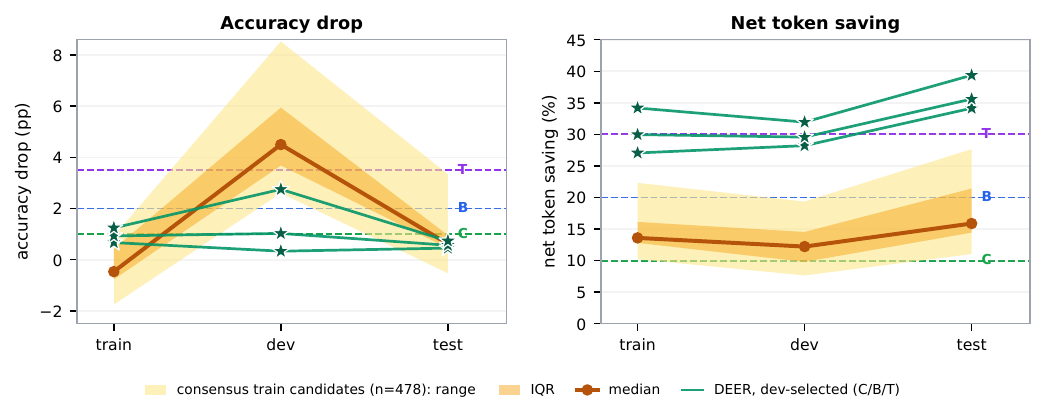}
  \caption{\textbf{A rule must clear the gate on every split.} The $478$ amber
  rules are those clearing the conservative gate \emph{in-sample on train},
  shown as median, interquartile range and full range. Dashed lines mark the
  three preregistered gates (\textbf{C}onservative, \textbf{B}alanced,
  \textbf{T}oken-efficient); a rule must sit below them on the left and above
  them on the right. On dev, the split the gate is applied on, none of the
  $478$ does; jointly, $0$ of $478$ clear all three splits, against $3$ of $3$
  for the dev-selected DEER points. Macro-averaged over $18$ environments per
  split.}
  \label{fig:split-transfer}
\end{figure*}

\section{Frontier and Reproduction Tables}
\label{sec:appendix-tables}

Figure~\ref{fig:ws-heatmap} traces the consensus trade-off along both sweep
axes; read down the window axis, the minimum attainable accuracy drop falls
monotonically (from $22\pp$ at $W{=}1$ to $0.8\pp$ at $W{=}30$) while the maximum
net saving falls with it (from $98\%$ to $39\%$), so the low-drop region is
reachable only where saving has already collapsed. Table~\ref{tab:deer} gives
the DEER threshold frontier that does reach the gate region.

\paragraph{Accuracy tax versus probe tax.}
Table~\ref{tab:grossnet} separates the two costs of \S\ref{sec:exp-sweep}:
under dense re-probing a late-stopping rule pays the probe output $p$ of
Section~\ref{sec:accounting} over a long prefix, opening a gap between its gross
savings $(B-s)/B$ and its net savings $(B-s-p)/B$. That gap is the probe tax. It
scales with probe density and is reducible, since sparser or shorter probes and
KV-cache reuse all shift the \emph{savings} axis. The \emph{accuracy} floor is
not reducible this way, because the ongoing correction behind it is present
wherever reasoning is still moving (\S\ref{sec:mech-ratio}). It is that floor,
not the probe tax, that keeps the safe-and-saving corner empty and that DEER
escapes.

\begin{table}[t]
  \centering\small
  \setlength{\tabcolsep}{3.4pt}
  \begin{tabular}{lcc}
    \toprule
    Consensus point & Gross saving & Net saving \\
                    & $(B-s)/B$    & $(B-s-p)/B$ \\
    \midrule
    drop $\le1.0\pp$ (safe rules) & $+2.07\%$ & $+0.21\%$ \\
    save$\ge10\%$ (drop $2.7\pp$) & $+14.9\%$ & $+10.9\%$ \\
    save$\ge20\%$ (drop $6.2\pp$) & $+27.0\%$ & $+20.2\%$ \\
    \bottomrule
  \end{tabular}
  \caption{Gross vs.\ net savings for the safe rules and two consensus frontier
  points; the gap ($\sim\!4$--$7\pp$) is the probe tax. The first row is the ceiling over
  the $40$ rules that clear the conservative accuracy cap: even untaxed they
  save $2.07\%$, far under the $10\%$ floor, so all three gates stay empty on
  gross saving as well.}
  \label{tab:grossnet}
\end{table}

\paragraph{The position--statistic ablation.}
Table~\ref{tab:ablation} varies those two factors one at a time; Table~\ref{tab:ablation-full}
gives the frontier behind it, the smallest accuracy drop with which each cell
buys a $10$, $20$ or $30\%$ saving. Under gross accounting, where the probe cost
is forgiven on both sides and exactly two factors remain, the gap between the
swept family and DEER decomposes additively into a statistic term and a
position term: $+2.60$ and $-0.64\pp$ at the $10\%$ floor, $+4.57$ and $-0.64\pp$
at $20\%$. The position term is negative, because at DEER's own signal our
denser grid is the slightly better place to read. The whole of the accuracy gap,
and more, is therefore the statistic. Under net accounting, where a cell's own probe cost
also decides which savings it can reach, the $20$ and $30\%$ floors split about
evenly between the two terms.
Binned into $5$-point savings bands, so that each cell is read at a saving it
actually delivers rather than one it overshoots, changing the positions lowers
the accuracy price in none of the eight bands at or above $10\%$ saving. Of the
nine bands in all, the one place it does help is the $5$--$10\%$ gross band,
$1.74$ to $0.81\pp$. Changing the statistic lowers the price in all nine.
Two limits on the table. The $30\%$ column understates the DEER
confidence signal at our positions: its fourteen thresholds jump from a $29.1\%$
to a $40.3\%$ gross saving, so at that floor it is charged $7.84\pp$ for savings
it was not asked to deliver. That same cell is also the one near miss in this
paper. At $13.9\%$ net saving and $+0.69\pp$ it satisfies both numeric legs of
the conservative gate and fails only its per-environment leg, saving tokens in
$14$ of $18$ environments against a floor of $80\%$; the four it fails are four
of the six DeepSeek AIME24 and AMC23 cells, six- and eight-problem sets on which
the dense probe schedule costs more than the stops it buys.

\begin{table*}[t]
  \centering
  \setlength{\tabcolsep}{11pt}
  \begin{tabular}{llrrrrrr}
    \toprule
    & & \multicolumn{3}{c}{Net} & \multicolumn{3}{c}{Gross} \\
    \cmidrule(lr){3-5}\cmidrule(l){6-8}
    Statistic & Reads at & $10$ & $20$ & $30$ & $10$ & $20$ & $30$ \\
    \midrule
    Agreement  & our grid   & $-2.7$ & $-6.2$ & $-11.8$ & $-1.9$ & $-3.9$ & $-8.1$ \\
    Agreement  & boundaries & $-3.8$ & $-10.2$ & $-13.5$ & $-2.7$ & $-9.6$ & $-13.5$ \\
    Confidence & our grid   & $+0.7$ & $-3.1$ & $-7.8$ & $+0.7$ & $+0.7$ & $-7.8$ \\
    Confidence & boundaries & $+0.1$ & $+0.1$ & $-2.8$ & $+0.1$ & $+0.1$ & $-0.4$ \\
    \bottomrule
  \end{tabular}
  \caption{The accuracy price of a saving, in points of macro accuracy: the
  best $\Delta$acc among rules reaching a $10$,
  $20$ or $30\%$ saving ($+$ a gain, $-$ a loss), for
  each combination of statistic and read positions, on the $659$ development
  problems of Table~\ref{tab:ablation}. Cells are searched over $3{,}520$ and
  $1{,}760$ preregistered rules respectively for agreement, and over DEER's own
  $14$ published thresholds for confidence; the coarser grid can only cost the
  confidence rows. Read down a column: changing the statistic lowers the price
  in all six, while changing the positions raises it at the $10\%$ floor under
  either accounting and at $20\%$ under gross. In the other three columns the
  position term also absorbs a probe tax (the dense grid probes $61.8$ times
  per problem against DEER's $8.6$ trials), and the confidence rows reverse.}
  \label{tab:ablation-full}
\end{table*}

\paragraph{Decision density.}
A third difference between the two families is how often each is allowed to
fire. On the development trajectories DEER's reasoning boundaries give it a
median of $10$ stop opportunities (mean $13.6$; its cap of $30$ binds on
$24.0\%$ of them), while the dense grid at interval $64$ gives a median of
$54$. More chances to fire is more chances to fire wrongly, whatever the signal
measures, so the schema varies the interval as one of its axes. At
$64/128/256/512$ tokens the cheapest rule in each block of $440$ loses
$3.52/1.91/1.69/1.57\pp$ for $11.18/7.82/1.26/1.65\%$ net saving, and no block
places a single rule inside the $1.0\pp$ cap; the $40$ rules that do are all
event-triggered, and the best saving among them is the $0.2\%$ of
\S\ref{sec:exp-sweep}. Interval $512$ is the
setting that matches DEER: a median of $7$ stop opportunities per trajectory
against DEER's $10$, and $6$ against $8$ on \textsc{MATH}500. There the
cheapest rule loses $1.57\pp$ for $1.65\%$ saving, worse on both axes than
DEER at its conservative point, $0.33\pp$ for $28.2\%$. Density is not what
separates them. Computed by \texttt{decision\_density.py}, which reproduces
the published $3{,}520/40/0.21\%$ population before it reports anything.

\paragraph{Probe prefill.}
Neither view charges the probe \emph{prompt} (\S\ref{sec:accounting}), and it is
by far the larger quantity: a probe re-reads the whole prefix, so on the fixed
schedules the prompt tokens come to $11$--$53\times$ the baseline's entire
decode budget, sparsest to densest, against $4.19\times$ for DEER at its
conservative point. Under the KV-cache reuse both \citet{dynasor_certaindex} and
\citet{deer} assume, that prefill is close to free; without it, it dominates
everything else. \citet{dong2026learnstop} report a policy that saves $32\%$ of
tokens under cache forking costing $121\%$ extra under black-box repeated
prefilling. The exclusion therefore favours the dense consensus schedules, not
DEER. Our accounting is otherwise the stricter one: scored under the
response-length metric of \citet{deer}, which forgives the discarded trial
output we charge, DEER's conservative point saves $30.3\%$ rather than the
$28.2\%$ we report, and the same gap holds at all $14$ thresholds
($+1.6$ to $+2.4\pp$).

\paragraph{Problem-pooled aggregation.}
Macro averaging gives the $8$-problem AMC23 cell the same weight as the
$100$-problem \textsc{MATH}500 one, so it is worth asking whether the empty
gate is a product of that choice. Table~\ref{tab:pooled-robustness} re-scores
the sweep with problems weighted equally and drops each environment in turn:
the gates stay empty throughout, but the margin by which they do is much
smaller once the small competition cells stop carrying equal weight.

\begin{table}[t]
  \centering\small
  \setlength{\tabcolsep}{3.4pt}
  \begin{tabular}{lcc}
    \toprule
    Weighting & Rules $\le1.0\pp$ & Best net saving \\
    \midrule
    Macro (protocol)     & $40$  & $0.21\%$ \\
    Pooled over problems & $634$ & $9.13\%$ \\
    \bottomrule
  \end{tabular}
  \caption{The dev sweep re-scored with problems weighted equally instead of
  environments. All three gates remain empty under either weighting
  ($0$ of $3{,}520$), and dropping any one of the $18$ environments leaves the
  outcome unchanged in all $18$ cases. Pooling is, however, far more permissive:
  the best rule that stays within the conservative accuracy cap saves $9.13\%$,
  $0.87\pp$ short of the $10\%$ floor, against $0.21\%$ under macro averaging.
  The protocol mandates macro averaging; pooling is a robustness check.}
  \label{tab:pooled-robustness}
\end{table}

\paragraph{Non-terminal stops along the whole window axis.}
Table~\ref{tab:flipdecomp} gives the measurement of \S\ref{sec:mech-ratio} at
every window, on all $3{,}420$ development-model trajectories. ``Differs'' is
the share of stops committing a non-terminal answer; ``harmful'' is the share
of the resulting accuracy changes that destroy a correct-in-the-end answer
(swaps excluded). Neither is compared against a chance baseline: a trajectory
that ends correct $93.3\%$ of the time has far more correctness to lose than to
gain, so some imbalance is implied by the accuracies alone, and the claim of
\S\ref{sec:mech-ratio} rests on the $155$ individual corrections observed
rather than on the size of the share. The last column is the wider scope of the
same count: truncated trajectories included and scored incorrect, so that a
correct stop on one counts as a rescue. It is the only place the direction
reverses, and only at $W{=}30$, where the rule fires on $7\%$ of problems and
saves $8\%$.

\begin{table}[t]
  \centering\small
  \setlength{\tabcolsep}{3pt}
  \begin{tabular}{rrrrrr}
    \toprule
    $W$ & Stops & Differs & Harmful & Net saving & Harm:rescue \\
        &       & (\%)    & (\%)    & (\%)       & (all stops) \\
    \midrule
    $1$  & $3167$ & $60.3$ & $99.0$ & $91.8$ & $1751{:}41$ \\
    $3$  & $3037$ & $33.7$ & $97.5$ & $75.9$ & $879{:}47$ \\
    $5$  & $2888$ & $20.5$ & $95.8$ & $59.7$ & $485{:}52$ \\
    $8$  & $2577$ & $13.7$ & $93.4$ & $43.6$ & $269{:}54$ \\
    $12$ & $2045$ & $10.6$ & $90.6$ & $32.0$ & $155{:}57$ \\
    $16$ & $1570$ & $9.0$  & $90.9$ & $23.5$ & $100{:}52$ \\
    $24$ & $887$  & $7.0$  & $93.3$ & $12.9$ & $42{:}40$ \\
    $30$ & $611$  & $7.2$  & $93.8$ & $8.2$  & $30{:}39$ \\
    \bottomrule
  \end{tabular}
  \caption{Non-terminal consensus stops along the window axis
  (\S\ref{sec:mech-ratio}). Widening the window drives ``differs'' down only to
  a floor near $7\%$, while the net saving that pays for it falls from $92\%$
  to $8\%$. Columns 2--4 are restricted to trajectories finishing inside the
  budget; the last column instead includes truncated ones, scored incorrect.}
  \label{tab:flipdecomp}
\end{table}

\paragraph{Prior-stopper reproductions.}
Table~\ref{tab:baselines} reports the two reproductions of
\S\ref{sec:exp-prior} in a common harness, on the same frozen trajectories and
under the same token accounting. \textbf{CertaIndex (CoT)} \citep{dynasor_certaindex} is
the consensus stop for a single chain of thought, reproduced from the released
implementation at its default \texttt{mid} setting (stop once three consecutive
probes give the same non-empty answer and none hedges), with the authors' probe
prompt and probe schedule. This is their rule as shipped, not a variant of ours.
\textbf{DEER} \citep{deer} here is the faithful readout reproduction (the
stronger trial-answer-submit variant is what \S\ref{sec:exp-deer} sweeps). Three
scope notes. First, CertaIndex's multi-path setting resembles
self-consistency, measuring entropy over answers from \emph{independently
sampled} trajectories, and nothing here bears on it; what we reproduce is the
single-trajectory case, where the same statistic is read over repeated probes
of one chain. Second, these are reproductions on our frozen trajectories and
benchmarks, not re-runs of either paper's end-to-end system, and the original
CoT results were reported on an easier benchmark, where intermediate answers
settle sooner. Third, the released implementation probes from position $0$
while we probe from $64$; from the third trigger on the two truncate at the
same position and deliver the same answer, and the only difference is that the
released code has one earlier opportunity to fire, which if taken stops it
sooner still, so the accuracy we report for it is an upper bound. Because a
single operating point is exactly what \S\ref{sec:exp-sweep} argues is
uninformative, we also report the other released \texttt{effort\_level}
settings in Table~\ref{tab:certaindex-effort}; the conclusion does not depend
on which one is chosen. We will release, but do not report, a prompt-adapted replay
of the same \texttt{mid} stop logic driven by \emph{our} \simplek{} probe
instead of the authors' answer suffix: it is the less aggressive of the two on
both axes, losing $44$--$66\pp$ while saving $71$--$89\%$, and it is not their
rule, so Table~\ref{tab:baselines} reports the faithful configuration.
Separately, the DEER reproduction here is plain DEER: DEER-Pro's confidence
calibration requires several varied answer-inducing prompts at the same
transition and the frozen bank stores one, so it is not computable offline
against these trajectories.

\begin{table*}[t]
  \centering\small
  \setlength{\tabcolsep}{4.2pt}
  \begin{tabular}{llccccc}
    \toprule
    Method & Model & Accuracy & $\Delta$acc & Fair token saving & Stop rate & Signal \\
    \midrule
    Full generation & \qweneight{}   & $85.4\%$ & --- & $0\%$ & $0\%$ & --- \\
    Full generation & \dsseven{} & $79.8\%$ & --- & $0\%$ & $0\%$ & --- \\
    \midrule
    CertaIndex (CoT) & \qweneight{}   & $15.3\%$ & $-70.1\pp$ & $90.1\%$ & $99.8\%$ & consensus \\
    CertaIndex (CoT) & \dsseven{} & $23.9\%$ & $-55.9\pp$ & $76.7\%$ & $98.7\%$ & consensus \\
    \midrule
    DEER       & \qweneight{}   & $86.2\%$ & $\mathbf{+0.8\pp}$ & $16.3\%$ & $41.4\%$ & boundary conf. \\
    DEER       & \dsseven{} & $74.9\%$ & $-4.8\pp$  & $20.2\%$ & $56.1\%$ & boundary conf. \\
    \bottomrule
  \end{tabular}
  \caption{Prior early-stoppers reproduced on our frozen trajectories (dev, macro
  over three benchmarks; probe tokens charged). \textbf{CertaIndex (CoT)} stops on
  almost every problem and stops early, and loses $56$--$70\pp$ while saving
  $77$--$90\%$ of tokens, the pattern \S\ref{sec:exp-fc} describes, at its
  default shipped operating point (Table~\ref{tab:certaindex-effort} sweeps the
  released alternatives). \textbf{DEER}, reading boundary confidence rather than
  consensus, stays near full accuracy ($+0.8\pp$ on \qweneight{}) while
  still saving tokens. These are reproductions on our benchmarks, not the original
  papers' end-to-end numbers.}
  \label{tab:baselines}
\end{table*}

\begin{table*}[t]
  \centering
  \setlength{\tabcolsep}{9pt}
  \begin{tabular}{lccccccc}
    \toprule
    & & \multicolumn{2}{c}{Train} & \multicolumn{2}{c}{Dev} & \multicolumn{2}{c}{Test} \\
    \cmidrule(lr){3-4}\cmidrule(lr){5-6}\cmidrule(lr){7-8}
    Effort level & Patience & $\Delta$acc & Saving & $\Delta$acc & Saving & $\Delta$acc & Saving \\
    \midrule
    \texttt{mild} & $8$ & $-18.1\pp$ & $59.4\%$ & $-31.6\pp$ & $53.9\%$ & $-16.6\pp$ & $66.7\%$ \\
    \texttt{low}  & $5$ & $-30.6\pp$ & $73.4\%$ & $-40.7\pp$ & $67.7\%$ & $-31.5\pp$ & $79.0\%$ \\
    \texttt{mid}  & $3$ & $-50.0\pp$ & $88.5\%$ & $-63.0\pp$ & $83.4\%$ & $-52.8\pp$ & $88.6\%$ \\
    \texttt{high} & $2$ & $-59.7\pp$ & $94.6\%$ & $-71.0\pp$ & $94.7\%$ & $-65.2\pp$ & $94.3\%$ \\
    \bottomrule
  \end{tabular}
  \caption{CertaIndex (CoT) across the released \texttt{effort\_level} settings
  (macro over the six model $\times$ benchmark cells as in
  Table~\ref{tab:baselines}, each pooling its three seeds; probe tokens
  charged). \emph{Patience} is
  the number of consecutive agreeing probes the setting requires. The
  \texttt{mid} row on dev is the setting reported in
  Table~\ref{tab:baselines} and reproduces it exactly. Accuracy falls
  monotonically as patience is relaxed, and the most lenient released setting
  still loses $31.6\pp$ on dev. A fifth released setting, \texttt{crazy},
  shares \texttt{high}'s patience of $2$ and differs only in probing every
  $32$ tokens rather than $64$; we collected no probe bank at that interval
  and do not reproduce it, and by the monotonicity above it cannot be more
  lenient than \texttt{high}.}
  \label{tab:certaindex-effort}
\end{table*}

\begin{table}[t]
  \centering\small
  \begin{tabular}{ccc}
    \toprule
    Threshold $\tau$ & Total drop (pp) & Net saving \\
    \midrule
    $0.9999$  & $-0.06$ & $20.8\%$ \\
    $0.999$   & $0.06$  & $25.4\%$ \\
    $0.995$   & $0.33$  & $28.2\%$ \\
    $0.99$    & $1.03$  & $29.6\%$ \\
    $0.97$    & $2.75$  & $31.9\%$ \\
    $0.95$    & $5.87$  & $34.8\%$ \\
    $0.90$    & $11.43$ & $42.6\%$ \\
    \bottomrule
  \end{tabular}
  \caption{DEER trial-answer-submit threshold frontier on dev (macro over $18$
  environments, same token accounting). Unlike consensus, low drop coincides with
  substantial saving; all three gates are
  cleared around $\tau\in[0.97,0.995]$.}
  \label{tab:deer}
\end{table}

\section{Supporting Figures: Sweep Surface, Selection, and Frontiers}
\label{sec:appendix-pareto}

\begin{figure}[t]
  \centering
  \includegraphics[width=\columnwidth]{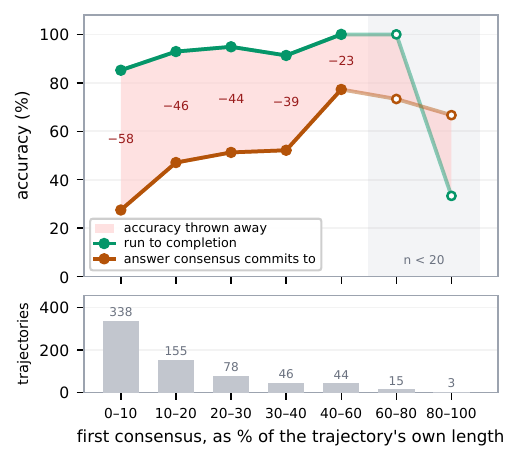}
  \caption{\textbf{Consensus forms early, where it is least reliable.} For each
  dev-split trajectory we take the first window of three agreeing non-empty
  probes and ask two things: is that answer correct, and would the same problem
  have been correct if left to run? Position is a fraction of the trajectory's
  \emph{own} length, so $500$ tokens into a $600$-token trajectory counts as
  late and $500$ into a $16$K one does not. Consensus forms in $679$ of $684$
  trajectories, half within the first tenth; the shaded gap is the accuracy a
  stop at that moment throws away. Open markers mark bins with fewer than $20$
  trajectories; the final bin ($n{=}3$) should not be read. Unlike the headline
  metrics elsewhere in the paper, the rates plotted here, and the ones quoted
  from them in \S\ref{sec:exp-fc}, pool the $684$ trajectories of the $18$
  development environments rather than macro-averaging over them, so that each
  position bin is read off the trajectories that actually fall in it.}
  \label{fig:consensus-pos}
\end{figure}

\begin{figure*}[t]
  \centering
  \includegraphics[width=\textwidth]{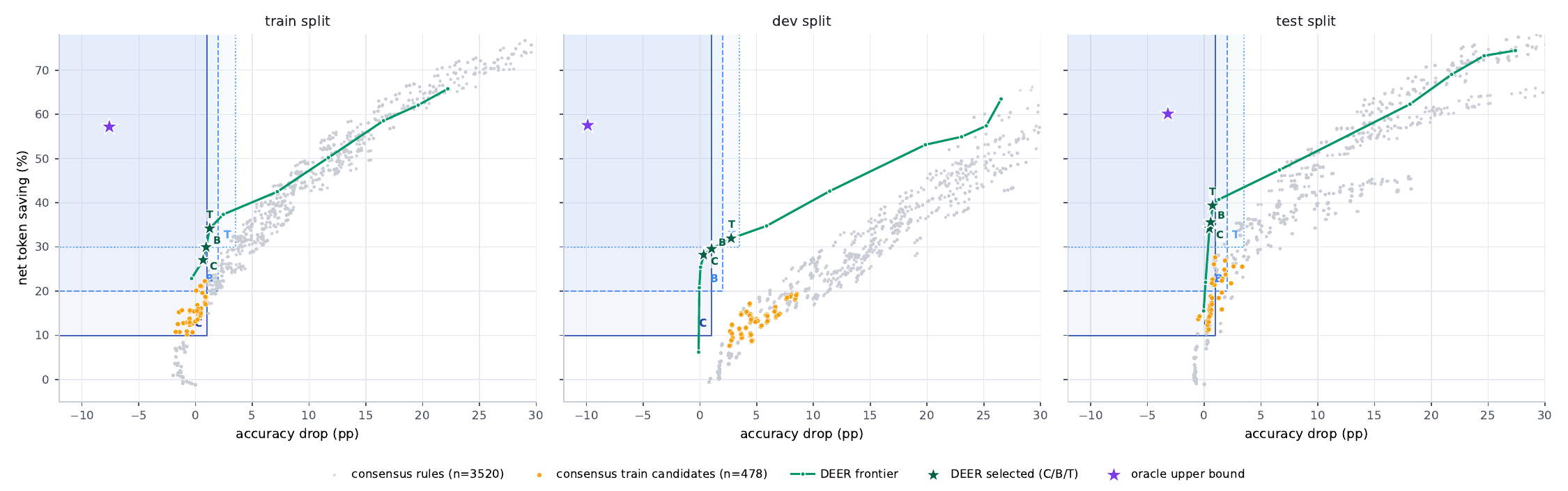}
  \caption{Selection $\rightarrow$ generalization across splits (two development
  models; macro-averaged over environments). Grey: all $3{,}520$ consensus
  rules. \textbf{Orange}: the $478$ rules that clear the conservative gate
  \emph{in-sample on train}. \textbf{Green}: the DEER threshold frontier, with
  the three dev-selected operating points
  (\textbf{C}onservative/\textbf{B}alanced/\textbf{T}oken-efficient) as stars.
  \textbf{Purple star}: the oracle upper bound (earliest correct probe).
  \textbf{Blue bands}: the three preregistered gates, darkest first. A rule
  lies inside a band when its drop is at most the cap and its saving at least
  the floor (\textbf{C} $1.0\pp/10\%$ solid, \textbf{B} $2.0\pp/20\%$ dashed,
  \textbf{T} $3.5\pp/30\%$ dotted).}
  \label{fig:app-splits}
\end{figure*}

\begin{figure*}[t]
  \centering
  \includegraphics[width=\textwidth]{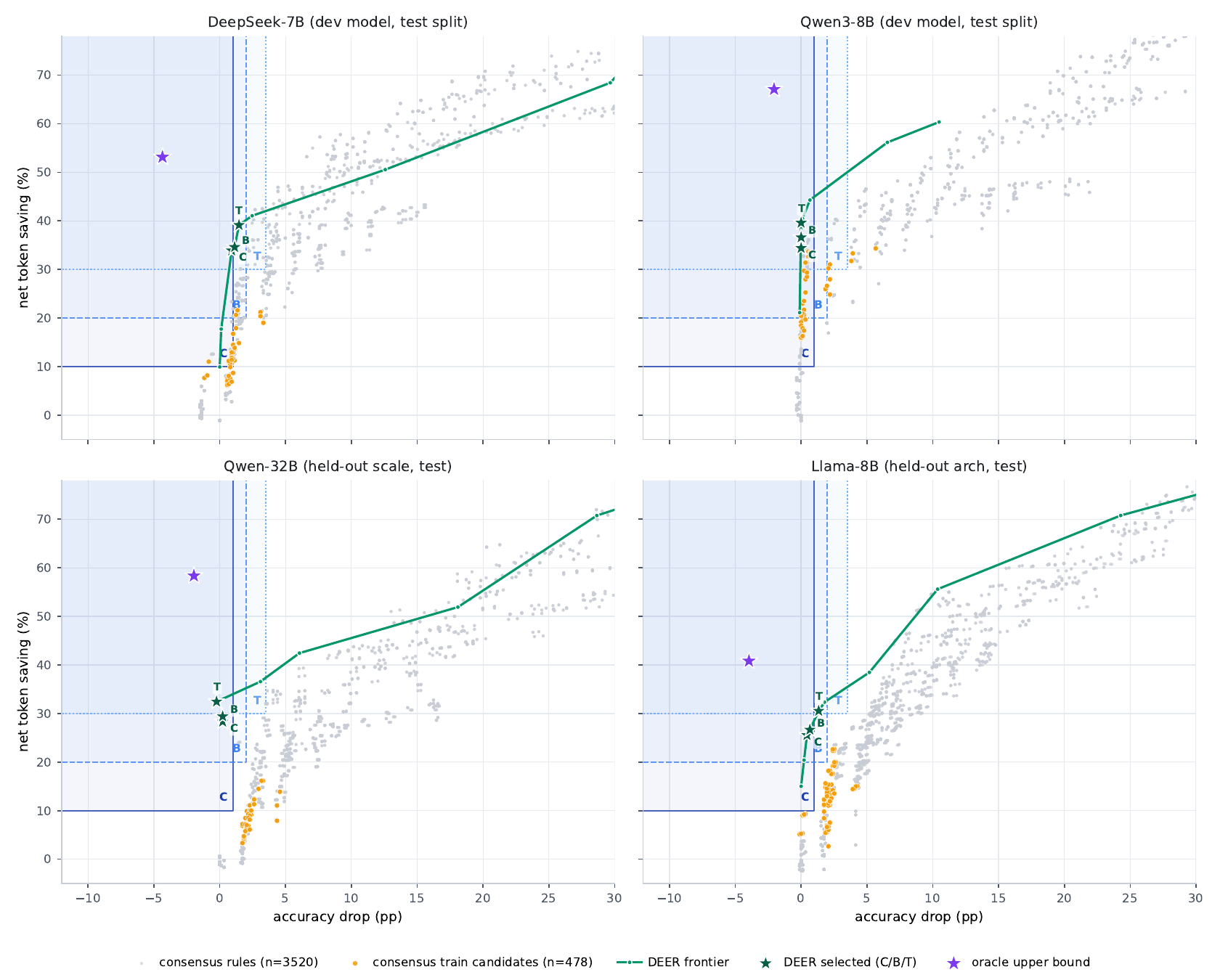}
  \caption{Out-of-sample generalization across scale and architecture (all panels
  test split). The two development models (top) are shown on their held-out test
  seeds; the two held-out models (bottom) are entirely unseen. Markers as in
  Figure~\ref{fig:app-splits}. The dev-selected DEER points stay inside the
  conservative gate on every model; no consensus rule clears it on all four. On
  the two development models many are admissible \emph{in-sample} on test, but
  none is selectable from dev (\S\ref{sec:exp-heldout}).}
  \label{fig:app-models}
\end{figure*}

\begin{figure*}[t]
  \centering
  \includegraphics[width=\textwidth]{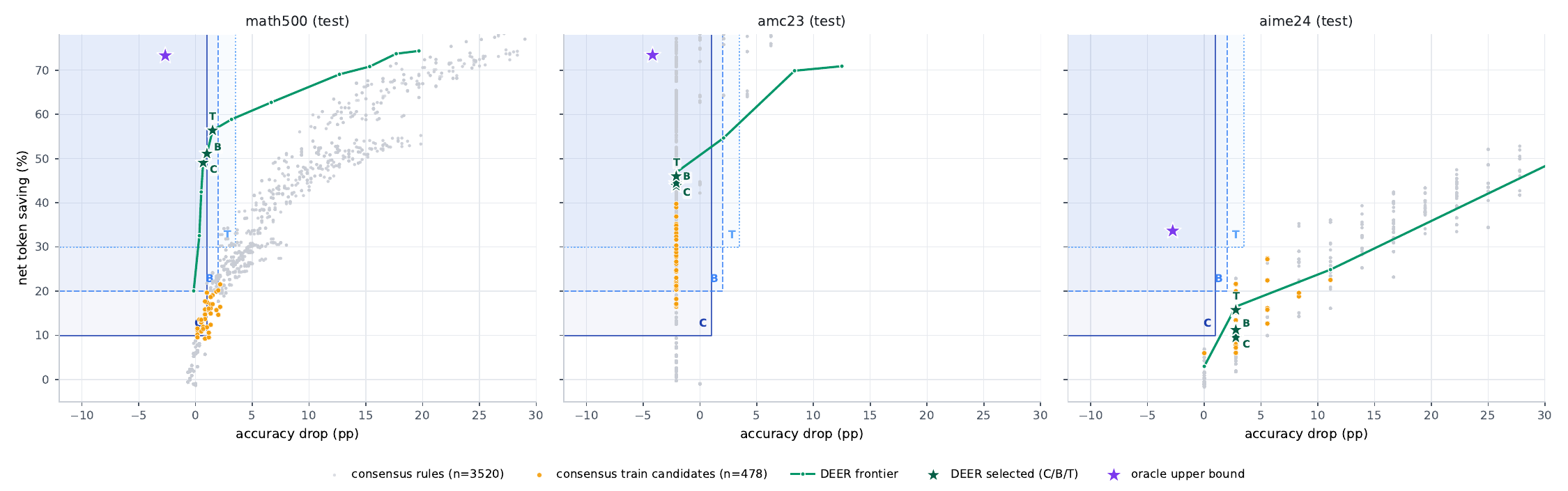}
  \caption{Out-of-sample generalization across benchmarks (test split,
  development models). The consensus trade-off and the DEER advantage hold on
  \textsc{MATH}500 and AMC23; on the small AIME24 set (few problems per
  environment) the two frontiers are noisier and intertwine. Markers as in
  Figure~\ref{fig:app-splits}.}
  \label{fig:app-bench}
\end{figure*}

This appendix collects the supporting figures referenced from the main text.
Figure~\ref{fig:ws-heatmap} reads the empty safe-and-saving corner directly off
the sweep's $(W,s)$ surface (\S\ref{sec:exp-sweep}); Figure~\ref{fig:split-transfer}
traces the train~$\rightarrow$~dev~$\rightarrow$~test selection pipeline
(\S\ref{sec:exp-heldout}); and Figure~\ref{fig:consensus-pos} shows where consensus
first forms as a fraction of each trajectory's length (\S\ref{sec:exp-fc}). Figures~\ref{fig:app-splits}--\ref{fig:app-bench} then
give the saving-versus-drop view of the same selection and generalization results,
resolved per split, per model and per benchmark, with the oracle upper bound on
each panel. They show the distributions the summaries are drawn from.

\paragraph{Selection-pipeline details (\S\ref{sec:exp-heldout}).}
In Figure~\ref{fig:split-transfer}, the $478$ consensus rules that clear the
conservative gate \emph{in-sample on train} are tracked onto dev and test as a
band; the three dev-selected DEER thresholds are tracked individually. On
dev, the split the gate is applied on, not one of the $478$ is admissible,
while all three DEER points stay inside on every split. The dev and test splits
are not equally demanding: the same $478$ rules have a median drop of $4.5\pp$
on dev but $0.6\pp$ on test, an asymmetry driven by the two smallest
environments (AMC23 and AIME24 under \qweneight{}).
The gate has to be read jointly rather than per split: $364$ of the
$478$ are admissible on test, but they are exactly the in-sample winners a
held-out split exists to reject, and none of them is selectable from dev.

\paragraph{Unseen-model details.}
On the $4\times$ larger \mbox{Distill-Qwen-32B} the conservative gate stays
empty (the best rule under a $1.0\pp$ drop saves $0.6\%$), but a scale effect
appears at the looser gates: a few consensus rules become admissible
\emph{in-sample} ($4$ at balanced, $6$ at token-efficient), reflecting that a
bigger model's intermediate answers stabilize somewhat earlier. These are not
the dev-selected rules, since dev admits none. On the same-scale
\mbox{Distill-Llama-8B} all three gates stay empty ($0/0/0$; the best rule
under a $1.0\pp$ drop saves $9.3\%$). The Llama-8B trajectories were
re-collected after correcting a chat-template defect (a missing model-specific
beginning-of-sequence token that degraded its generations); the corrected model
scores $\sim\!88\%$ on \textsc{MATH}500. The \emph{oracle} upper bound on each
panel is a perfect probe-based stop committing the earliest correct probe; it
sits at a negative drop of $2$--$5\pp$ at $40$--$80\%$ saving, far in a corner
neither method reaches.

\section{Case Studies}
\label{sec:appendix-cases}

The mechanism of Section~\ref{sec:mechanism} is easiest to see on individual
trajectories. All three below are \dsseven{} on \textsc{MATH}500, seed $42$,
probed every $64$ generated tokens; we write $a\times n$ for an answer
repeated over $n$ consecutive probes. A rule requiring three agreeing probes
fires at the third element of the first such run.

\paragraph{Case 1: a persistent placeholder.}
Problem~$320$ asks for $\cos\angle RPS$ given $\sin\angle RPQ=\tfrac{7}{25}$
(answer $-\tfrac{24}{25}$). The probe stream is
\begin{center}\small
\texttt{0}$\times$\texttt{27},\; \texttt{B}$\times$\texttt{8},\;
\texttt{-24/25},\; \texttt{B}$\times$\texttt{7},\\
\texttt{E},\; \texttt{B}$\times$\texttt{6},\;
\texttt{-24/25}$\times$\texttt{7}
\end{center}
The model answers \texttt{0} at the very first probe and repeats it for $27$
consecutive probes, covering tokens $64$--$1728$ of a $3{,}683$-token
trajectory. It then passes through a stretch of bare option letters before
reaching $-\tfrac{24}{25}$ and holding it to the end; the trajectory is correct.

This single case makes three points. The stop is not a considered wrong answer
but a value emitted before any real work has been done, and it is perfectly
stable for nearly half the trajectory. That it never changes therefore says nothing about
whether reasoning has finished. And because the run is $27$ probes long, it
defeats \emph{every} window in our sweep: even $W{=}24$, the second-largest we
searched, fires here and commits \texttt{0}. Enlarging the window does not
address a placeholder that persists longer than the window.

\paragraph{Case 2: a probe-format artifact.}
Problem~$253$ is ``What fraction of 2 feet is 3 inches?'', a free-response
question with no answer options. The stream is
\begin{center}\small
\texttt{3},\; \texttt{B},\; \texttt{3},\; \texttt{24},\; \texttt{B},\;
\texttt{D}$\times$\texttt{20},\; \texttt{1/8}
\end{center}
From token $384$ to token $1600$, more than two thirds of the trajectory, every
probe returns the letter \texttt{D}. There is nothing for \texttt{D} to refer to.
The model is not reporting a belief about the answer; it is completing the probe
suffix in a plausible format while its reasoning continues underneath. The
correct value $\tfrac{1}{8}$ appears only at the final probe. Agreement here is
maximal and perfectly uninformative, which is why any deployed rule needs to
check answer \emph{shape} before treating agreement as evidence
(\S\ref{sec:mech-placeholder}).

\paragraph{Case 3: a wrong answer either way.}
The minority case is equally worth stating. Problem~$240$ (gold $116$) gives
\begin{center}\small
\texttt{52}$\times$\texttt{2},\; \texttt{104},\; \texttt{52}$\times$\texttt{5},\;
\ldots,\; \texttt{154}$\times$\texttt{8},\; \texttt{52},\;
\texttt{154}$\times$\texttt{62}
\end{center}
A three-probe rule fires at token $384$ and commits \texttt{52}; the trajectory
itself ends at \texttt{154}. Both are wrong. Here early stopping destroys
nothing, because continued reasoning was not going to arrive at the right answer
either; the model settles firmly on \texttt{154} and holds it for its final $62$
probes. Cases like this are why \S\ref{sec:mech-ratio} keeps stops that swap one
wrong answer for another separate from those that \emph{change}
correctness: on problems the model was never going to solve, a consensus stop
costs accuracy nothing and saves tokens, which is the regime consensus early exit
was designed for. The difficulty is that agreement does not distinguish this case
from Case~1, and Case~1 is the more common one.

%% file: G_self_consensus.tex
\section{Self-Consensus: Full Analysis}
\label{sec:appendix-fc}

Section~\ref{sec:exp-fc} summarizes what this appendix
establishes in full: in an intervention-free setting, self-consensus is not a
reliable stopping signal.

\subsection{Setup}
\label{sec:fc-setup}
We analyze the frozen main trajectories (Section~\ref{sec:method}): \dsseven{} on
all $500$ \textsc{MATH}500 problems \citep{lightman2024verify,hendrycks2021math}
at the main $16$K-token budget, temperature $0.6$, top-$p$ $0.95$, across three
seeds ($1{,}500$ trajectories in total).\footnote{Train/dev problem ids are
collected at seeds $42/43/44$ and the held-out test ids at seeds $45/46/47$;
their union covers all $500$ problems. This is a descriptive characterization of
the frozen trajectories (it selects and tunes nothing), so it does not touch the
preregistered sweep or the test-split commitment.} The dense probe bank appends
the boxed-answer probe suffix every $64$ generated tokens and reads off the
model's current answer. The probe is a separate forward pass: the controller only
\emph{observes}, never halting or altering the main trajectory. This yields, per
trajectory, the full answer sequence a stopping rule would have access to at the
budget used throughout the paper.

\subsection{Agreement at the End of a Trajectory}
Agreement measured at the \emph{end} of a trajectory does track correctness.
On the $186$ trajectories that reach unanimous \emph{cumulative} agreement (all
non-empty probes concur), $97.8\%$ are correct, with only four agreeing throughout on
a wrong answer; and of the $1{,}205$ whose \emph{final five-probe window} is
unanimous, $90.4\%$ are correct. But this end-state agreement is not what an
online stopping rule can act on: a rule fires the \emph{first} time it sees
agreement, mid-trajectory, long before the answer has settled. The rest of this
appendix shows that \emph{early} agreement, the signal a controller must
actually use, is far from terminal.

\subsection{A Naive Consecutive-Agreement Stop}
We simulate a \emph{naive consecutive-agreement} stop: halt at the first point
where three consecutive probes agree, are non-empty, and are \emph{certain}
(the model emits the boxed answer with high token-level confidence; formal
definition in Appendix~\ref{sec:appendix-schema}). This is a simplified
heuristic, \emph{not} the CertaIndex (CoT) rule, which we reproduce from its
released implementation in
\S\ref{sec:exp-prior}. It fires on $1{,}477/1{,}500$ trajectories. The committed
answers are correct only $50.5\%$ of the time, whereas letting those same
problems run to completion reaches $90.7\%$, a
$40.2\pp$ accuracy loss in exchange for an average saving of $3301$
tokens, with $731$ trajectories stopped on an outright wrong answer.

\subsection{Recovery After an Early Consensus}
The loss is driven by \emph{recovery}: continued reasoning overturns the
intermediate answer far more often than intuition suggests. Of the $736$
trajectories that at some point formed a three-probe consensus \emph{different}
from their eventual final answer, $620$ ($84.2\%$) ultimately ended with a
correct \emph{final} answer.

Normalizing by position \emph{within each trajectory} shows where the damage
concentrates: reliability \emph{improves} the later the first consensus falls in
a trajectory's own span, so the earliest consensus, the one an online rule acts
on, is the least reliable (\S\ref{sec:exp-fc},
Figure~\ref{fig:consensus-pos}).

%% file: H_practice.tex
\section{Using the Negative Result}
\label{sec:appendix-practice}

The gates are empty, but the measurements behind them are constructive about
what a stopping signal has to do.

\paragraph{Agreement as a precondition.}
Agreement is cheap to read and, once a trajectory has finished, does track
correctness (\S\ref{sec:exp-fc}); what it cannot do alone is carry the stop. A
probe does not report a belief; it \emph{requires} an answer, and early in a
trajectory what it returns is substantially a property of the query, since two
differently worded probes read back different answers $54\%$ of the time in the
first tenth (\S\ref{sec:mech-wording}). Agreement over such readings
establishes that the elicited answer has not moved, not that the reasoning has
terminated. It can serve as a necessary precondition, with the decision left to
a signal that separates a settled trajectory from one still working.

\paragraph{Measuring convergence.}
Demanding more agreement buys safety only by giving the saving back
(\S\ref{sec:mech-ratio}), so the useful direction is a better reading of
whether the reasoning has converged. Two candidates follow from our measurements, and we evaluate neither:
requiring two differently worded probes to agree, which
\S\ref{sec:mech-wording} shows is demanding early and easy late; and comparing
the \emph{reasoning} across positions, the semantic similarity between
successive prefixes, instead of the low-bandwidth boxed answer span.

\paragraph{Transfer across splits.}
Of our rules, $478$ clear the conservative gate \emph{in-sample on train} and
$364$ are admissible on test, so a study reporting on one split
would have found a working consensus rule (\S\ref{sec:exp-heldout}). Not
one is selectable from dev, and none clears the gate on all three splits
jointly: one environment or one split is enough to produce a rule that looks
both safe and saving.